\documentclass[11pt,letterpaper]{article}
\usepackage{cogsys}
\usepackage[T1]{fontenc}
\usepackage{times}
\usepackage[pdftex]{graphicx} 

\usepackage{natbib}
\cogsysheading{11}{2021}{1--**}{9/2021}{11/2021}

\ShortHeadings{A System for Image Understanding using Sensemaking and Narrative}
              {Z.\ Battad and M.\ Si}

\begin{document} 

\title{A System for Image Understanding using Sensemaking and Narrative}
 
\author{Zev Battad}{battaz@rpi.edu}
\author{Mei Si}{sim@rpi.edu}
\address{Department of Cognitive Science, Rensselaer Polytechnic Institute, 
         Troy, NY 12180 USA}
\vskip 0.2in
 
\begin{abstract}
Sensemaking and narrative are two inherently interconnected concepts about how people understand the world around them. Sensemaking is the process by which people structure and interconnect the information they encounter in the world with the knowledge and inferences they have made in the past. Narratives are important constructs that people use sensemaking to create; ones that reflect provide a more holistic account of the world than the information within any given narrative is able to alone. Both are important to how human beings parse the world, and both would be valuable for a computational system attempting to do the same. In this paper, we discuss theories of sensemaking and narrative with respect to how people build an understanding of the world based on the information they encounter, as well as the links between the fields of sensemaking and narrative research. We highlight a specific computational task, visual storytelling, whose solutions we believe can be enhanced by employing a sensemaking and narrative component. We then describe our system for visual storytelling using sensemaking and narrative and discuss examples from its current implementation. 
\end{abstract}

\section{Introduction} 
 
\begin{figure}[t]
\vskip 0.05in
\begin{center}
\includegraphics[width=0.63\textwidth]{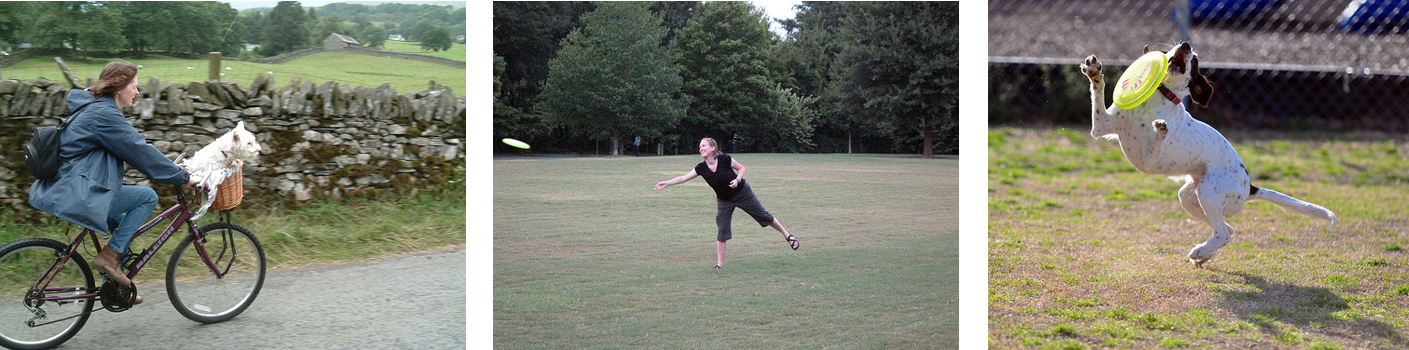}
\caption
{A sequence of scenes that a person might encounter over the course of a regular day.}
\label{figure:set-17-images}
\end{center}
\vskip -0.2in
\end{figure}

When humans navigate the world, they are bombarded with a plethora of information - sights, sounds, scents, textures. This cascade of sensory information gives a person direct information about the world around them. However, people do not solely understand the world as a cascade of sensory information. Rather, people glean a higher level of meaning - that a cluster of shapes and colors and sounds is a dog, that the dog is catching a frisbee, that a woman is throwing a frisbee and smiling, etc (Figure \ref{figure:set-17-images}). These direct observations represent a structuring and interconnection of lower levels of information. However, people also do not solely understand the world as a cascade of observations. Rather, from a cascade of observations, a person imagines a conceptual expanse outside of the directly observable to help tie together a holistic account of the information they encounter - that the frisbee observed in two separate perceptual instances is actually the same frisbee; that the woman throwing that same frisbee leads to the dog catching it because that is how playing frisbee typically proceeds; that the woman is smiling because she is having fun playing, but might be tired and dirty and want to get clean soon, etc.

The process that human beings use to structure and interconnect information is called sensemaking. Organizational theorist Karl Weick characterizes sensemaking as a continuous process, a "continued redrafting of an emerging story so that it becomes more comprehensive, incorporates more of the observed data, and is more resilient in the face of criticism" \citep{weick2005organizing}. The "redrafting" process that Weick alludes to is speculation towards, and formation of, an explanation about how the world works and how information is related to each other. Sensemaking has also been cast as a process for creating a representation of information to help answer specific questions about domains of information \citep{russell1993cost,russell2009overview,kolko2010sensemaking}, as well as a method for continuously redressing a network of connections between facts \citep{klein2006making1,klein2006making2}. Theories of sensemaking agree that sensemaking is about interconnecting information, elaborating with additional conclusions outside of the information itself, and structuring information to help a person better understand the world. 

One of the crucial end-points of sensemaking is the human construct of narrative. A narrative is a connected sequence of events with objects, people, and places - often referred to as \textit{existents}. Narratives are reflective of the connections humans use to construct an account of the world they encounter. This is why the markers of narrative coherence - temporal and spatial orientation, causal ordering and continuity, and thematic and emotional evaluations - appear so regularly in natural human-made narratives \citep{givon1992grammar,habermas2000getting}. It is also why the coherence of personal narratives is a reliable measure of mental health; the ability to tie together a sensible narrative account of the events in one's life is considered a critical and regular skill for healthy human beings \citep{reese2011coherence}. At a more fundamental level, narrative is thought to be inherent to the way people represent what they know of the world. Narrative has been posited as intrinsically linked to how humans organize knowledge, link their experiences and memories, and cement their understanding of time \citep{neumann2008introduction,bruner2001self,abbott2008cambridge}. 

Taken together, sensemaking provides a process for interconnecting and elaborating on information encountered in the environment to help a person create an understanding of the world. Narrative provides one form of that understanding, including specific types of coherent connections between pieces of information. We believe that this combination of methods, which is successful and ubiquitous for humans trying to parse the world, can be applied to improve computational systems attempting to do the same.

In this paper, we will discuss our computational system for visual storytelling. The system generates interconnected knowledge graphs from the observed information in a sequence of images. We will introduce the theories of sensemaking and narrative connections that inform the system's design, the computational sensemaking process the system uses to hypothesize narrative connections between pieces of observed information, and examples of its current implementation. 

\section{Background}
\label{section:background}

Two relevant areas of literature we review for this work are theories for human sensemaking from cognitive science and the coherent connections used in narrative from studies in narrative. For sensemaking, we identify three key aspects of sensemaking which inform the design of our system's computational sensemaking process. We will then outline the categorization used for our system for the types of connections used to coherently tie together narratives. 

\subsection{Sensemaking}
\label{section:sensemaking}

Sensemaking is the process of creating consistency and coherence between observations in the environment and an existing understanding of the world, in so doing expanding one's understanding of both. There are three theories of sensemaking that are most relevant to the approach in this paper: Weick's model, Klein, Hoffman \& Moon's model, and Russel's model. 

Weick's model for sensemaking explains sensemaking as the cyclic construction and restoration of plausibility in the face of new information in the environment. An important part of the process is the ongoing updating of a person's understanding of a situation according to information in the environment \citep{weick1995sensemaking}. Weick stresses that the process of sensemaking is more driven by plausibility than accuracy. As Weick says, "sensemaking is not about truth and getting it right. Instead, it is about continued redrafting of an emerging story so that it becomes more comprehensive, incorporates more of the observed data, and is more resilient in the face of criticism" \citep{weick2005organizing}. People seek explanations even if they do not have the most accurate idea of the information encountered. In these cases, the best strategy is to integrate the information encountered as best as one can with prior knowledge and prior inferred plausible explanations. 

Klein, Hoffman, and Moon's theory of sensemaking also functions on building mental models of information in the environment. Their work places more emphasis on the connections between information and is oriented towards making sense of data rather than situations. In their theory, sensemaking is done for future decision-making based on past information, constructing an understanding of what has happened in the world in the process. Klein, Hoffman, and Moon cast sensemaking as "a motivated, continuous effort to understand connections (which can be among people, places, and events) in order to anticipate their trajectories \citep{klein2006making1}." The connections between people, places, and events, all three of which are fundamental building blocks of narrative, are especially relevant to our current work \citep{klein2006making2}. 

Russel, who like Klein, Hoffman, and Moon also focuses on making sense of data, offers a task-oriented model for sensemaking. In Russel’s conception, a crucial part of sensemaking is the creation of a representation of the information encountered: “the formation of a model... which can then be used as a hypothesis upon which to examine, test, and accept or reject specific questions and ideas” \citep{kolko2010sensemaking,russell1993cost}. The representation built from a set of information during the sensemaking process helps people ask and answer questions to themselves about that information, even as the representation is being built. 

The three models of sensemaking reviewed above agree on three key aspects of sensemaking: 

\paragraph{Interconnecting information.} In the theories reviewed above, interconnected information is either existing knowledge/prior conclusions about the world, as in Weick's model; any piece of information with any other piece of information, as in Klein, Hoffman, and Moon's model; or competing/parallel sets of information, as in Russel's model. Whatever the specific pieces of information are, their interconnection is important.
\paragraph{Elaborating upon information with conclusions outside of the information itself.} In Weick's and Klein, Hoffman, and Moon's models, elaboration is done either with existing knowledge or with new information sought out on subsequent iterations of the sensemaking and information gathering process. Though Russel de-emphasizes the role of global knowledge, his model also includes new conclusions about existing parallel pieces of task-specific information.
\paragraph{Reaching a more internally consistent understanding of the world.} In all three models, sensemaking is a process that makes information more consistent with itself and with other connected pieces of information, more internally self-confident, and better structured to help draw conclusions that a human being would find sensible about the world. 

\paragraph{}
These aspects of sensemaking inform the design of our system's computational sensemaking process. The system aims to interconnect information by hypothesizing narrative connections between observed pieces of information. It does so by drawing on evidence from its own observations and from outside of its observations: from external knowledge and its own prior hypotheses. Finally, the hypotheses it creates are selected and kept internally consistent through a multi-objective optimization process. 

\subsection{Narrative Connections}
\label{section:narrative_connections}

In the definitions of sensemaking reviewed above, there is no true end-point for sensemaking. It is a continuous process that humans engage in as they live their lives and interpret the world. In Weick's words, it is a "continued redrafting of an emerging story" \citep{weick2005organizing}. This does not make intermediary accounts about specific sets of information - discrete drafts of the emerging story - any less valid. Humans regularly share finite experiences and understandings, bundling them into accounts that they can communicate with one another, i.e. narratives. 

The connections that tie together an individual's narrative account are the elements of narrative coherence. The elements of narrative coherence found regularly in human-made narratives inform the types of connections, and thus the hypotheses, a person might speculate when expanding their understanding of the story behind a set of images. More generally, elements of narrative coherence are reflective of the things humans place importance in when tying together an account of the world they encounter. 

Though all narratives contain these elements of narrative coherence, the type of narrative most salient to every day sensemaking is the personal narrative, i.e. a story from or about one's personal or every day life. Reese et al. explores the connections that patients make in personal narratives in the context of clinical psychology \citep{reese2011coherence}. There, a patient's ability to consistently create well-structured narratives about their lives is related to one of several markers for psychological well being. Reese et al. categorizes the well-structuredness of personal narratives into a clinical tool called the Narrative Coherence Coding Scheme (NaCCS). In the NaCCS, three major categories are used: context, chronology, and theme. Context encompasses spatial and temporal orientation, e.g. time and place of events. Chronology is the ordering of actions, closely related to the temporal aspect of context. The final category, theme, encompasses causal links, personal emotional and motivational evaluations, and connections to previous events in a person's life. 

The first two categories of Reese's coding scheme are similar to Gernsbacher \& Givon's categorization of grammatical markers of coherence in human written text: spatial, temporal, and referential continuity \citep{givon1992grammar,gernsbacher1995coherence}. The last category includes more comprehensive coherent links than the first two, and is similar to the categories found by Rideout in personal narratives delivered in court cases. Rideout observed causal and character consistency as major elements of narrative coherence, including consistency of character motivations and causal links between events \citep{rideout2013twice}. Consistency of character motivations relates more broadly to general character consistency and believability, which have been pointed out as major components of people's feelings of coherence in narratives \citep{riedl2010narrative,rapp2001readers}.

From the studies above, major elements of narrative coherence can be categorized as \emph{spatial, temporal, causal, referential, and affective (emotional/motivational)}. Though character consistency appears implicitly from the connections to and from people or animals, it is a shallow model of characters' emotions, motivations, and relationships relative to explicit character modeling done to enforce character consistency. Each element of narrative coherence represents a categorization of the different types of connections that a person can place between pieces of information in a narrative. As such, they also represent the types of connections a person can place between pieces of information through the sensemaking process, and the types of connections our system places between pieces of information in its computational sensemaking process. 

\section{Approach}

Human beings use the process of sensemaking to form an understanding of the world around them. One of the important structures of that understanding are the narrative coherence connections they place between pieces of information. We believe that a computational system with a similar goal of forming an understanding of the world would benefit from a similar sensemaking and narrative component. In this section, we will highlight our specific computational task and describe the architecture for our current system for accomplishing one part of that task. 

\subsection{Visual Storytelling}
\label{section:visual_storytelling}

The specific computational task we target for our approach is the Visual Storytelling task. The Visual Storytelling task is the task of computationally generating narrative text from sequences of images \citep{huang2016visual}. Various neural network-based approaches have been explored for automatically constructing stories for the Visual Storytelling task \citep{wang2018show,kim2018glac,huang2019hierarchically}. These methods for Visual Storytelling focus on capturing the directly observable. They do not focus on explicitly determining and representing what the story behind the images is beyond the directly observable, as humans do. When humans write stories for the visual storytelling task, they include a conceptual expanse outside of, but related to, that which they observe, with emotions, motivations, causal chains, etc. To reach these narrative evaluations beyond the directly observable, we believe a computational system that uses sensemaking and narrative coherence connections would be beneficial. 

\subsection{System Architecture}

\begin{figure}
    \centering
    \includegraphics[width=0.75\textwidth]{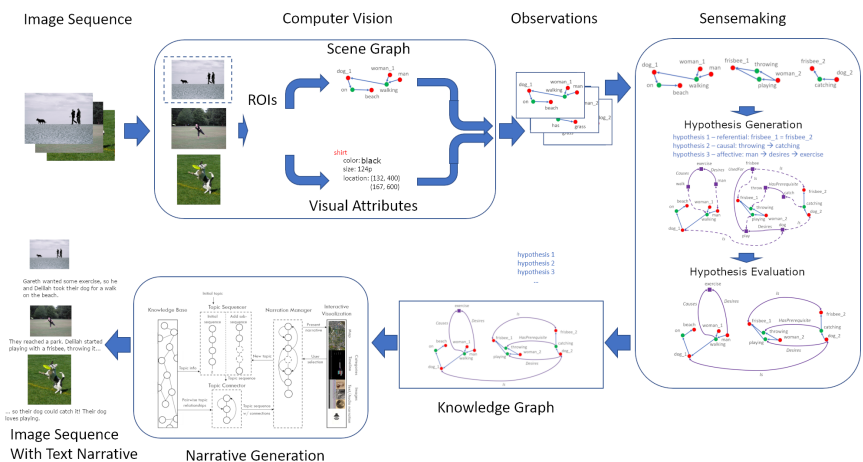}
    \caption{Architecture for visual storytelling system.}
    \label{figure:architecture}
\end{figure}

We are creating a system to generate text narratives from image sequences through a computational sensemaking process using narrative coherence connections. The architecture for the full system can be seen in Figure \ref{figure:architecture}. Our current work focuses on the Sensemaking subsystem. The goal of the subsystem is to create an intermediary knowledge graph output from visual information in a sequence of images and narrative coherence connections between those pieces of information. The examples in section \ref{section:examples} are generated from our current implementation of the sensemaking subsystem. 

\subsection{Computer Vision Subsystem}
\label{section:computer-vision-subsystem}

The Computer Vision subsystem handles observable information about the image sequence. The input to the Computer Vision subsystem is a sequence of images. The output is a set of observations about the images in the form of a scene graph. Scene graphs are a semantic representation of the regions of interest (representing objects, characters, and locations) in an image and the relationships between them, including any events that might be taking place. Following narrative terminology, we will refer to these objects, characters, and locations in regions of interest as \textit{existents}. The result is a set of existent-relationship-existent triples which, when connected, form a graph. 

Scene graphs serve as the observational information for the sensemaking subsystem. Though there are existing scene graph generation models, the current system leverages the Visual Genome dataset, a large repository of human-made scene graphs. Visual Genome is a dataset of 108,077 images with four types of human annotations: region descriptions, attributes (e.g. color, size), relationships, and question-answer pairs \citep{krishna2017visual}. The region descriptions and relationships form scene graphs for each image, which are passed as input to the sensemaking subsystem. 

\subsection{Sensemaking Subsystem}

In the system, observational information is represented by scene graphs of image sequences. To interconnect this information, the system hypothesizes narrative coherence connections, drawing from observations within the images, existing knowledge outside of the images, and its own prior hypothesized connections. This interconnection is constrained by the concern of remaining internally consistent and self-confident. 

The system uses a two-step process. First, hypotheses are generated by drawing from observations and commonsense knowledge sources, which serve as the system's source of existing knowledge about the world, and from its own prior hypotheses (Section \ref{section:hypothesis_generation}). The over-generated set of hypotheses is then evaluated, and a reduced version is accepted, obeying constraints derived from human sensemaking (Section \ref{section:hypothesis_evaluation}).

\subsubsection{Hypothesis Generation Process}
\label{section:hypothesis_generation}

For hypothesis generation, the system hypothesizes what narrative coherence relationships to place on and between observed pieces of information. These are the conclusions outside of the observed information that the system elaborates during sensemaking. Each hypothesis is premised by drawing from a combination of observational evidence, knowledge evidence, and prior hypotheses. Each type of evidence has a corresponding score, representing how confident the system is in the piece of evidence. 

Observational evidence comes from of the scene graph directly. Currently, they consist of attributes found on scene graph nodes, such as color. The score of the evidence depends on the computer vision subsystem's confidence rating of its observations. As a proxy for this, for Visual Genome annotations, observational evidence is given a score equal to the number of annotators that include it in the scene graph. 

Existing knowledge evidences comes from commonsense knowledge sources. The system uses ConceptNet as its source of commonsense knowledge. ConceptNet is a crowd-sourced network of commonsense information, with concepts acting as nodes and relationships between concepts acting as edges \citep{speer2017conceptnet}. ConceptNet uses a finite set of relationships for its edges. We categorize the relationships in ConceptNet into the five types of narrative coherence relationships identified prior for use in the system. The system links concept nodes in ConceptNet to corresponding scene graph nodes and traverses the edges between concept nodes according to this categorization to provide commonsense evidence for narrative coherence relationships between scene graph nodes. Weights from ConceptNet determine the score of existing knowledge evidence evidence. 

Just as sensemaking can draw from prior conclusions that a sensemaker has made about a specific situation, the hypotheses in the sensemaking subsystem can also be premised on other hypotheses. A hypothesis adds the evidence score of any hypothesis it is premised on to its own. 

Hypothesis generation for connections from three types of narrative coherence relationships are currently implemented: referential, causal, and affective. Referential and causal were chosen to tie together scene graph nodes across images - referential relationships connect existents, causal relationships connect events. Affective was chosen to provide more compelling examples of human-like evaluations about situations. Spatial and temporal relationships, which are more closely related to observational information, are left as future work. 

\paragraph{Referential Hypothesis Generation}

Referential relationships concern what existents across different images represent the exact same existent (e.g. a single person appearing across multiple images) or are parts of the same existent (e.g. the articles of clothing on a person or the parts of a bike). Referential relationships are an important foundation for other types of coherence relationships. The system currently implements hypothesis generation for a single type of referential relationship, the \textit{is} relationship, denoting which scene graph nodes represent separately encountered instances of the exact same existent. 

Referential hypothesis generation is premised on observational evidence and existing knowledge evidence. Observational evidence comes from matching scene graph nodes' attributes. Each matching attribute contributes an evidence score of 1. Existing knowledge evidence is found by checking the ConceptNet concept nodes corresponding to each scene graph node for matching concepts. The 0-length path through ConceptNet between two existents that have the same concept contributes an evidence score of 1. An example of referential hypotheses and its evidence can be seen in Figure \ref{figure:set-17-generation-examples}a. 

\paragraph{Causal Hypothesis Generation}

Causal relationships concern which events lead to or from other events via some underlying causal purpose or connection (e.g. putting on a jacket before walking into a snow-storm), establishing coherent lines of action in a narrative. The system currently implements hypothesis generation for a single type of causal relationship, the \textit{sequence} relationship, denoting which scene graph nodes represents events that are related enough to be considered lying upon the same coherent line of action. 

Causal hypothesis generation is premised on existing knowledge evidence and existing hypotheses. Existing knowledge evidence is found by searching for paths between two events' concept nodes using ConceptNet relationships categorized as causal, contributing an evidence score equal to the average weight of the edges along the path. The 0-length path between two events that have the same concept (e.g. two instances of people throwing something) is included in this definition and contributes an evidence score of 1. Existing hypotheses are found by comparing the existents that take part in each event in the scene graph. If any pair of existents have a referential \textit{is} hypothesis linking them, it means the same existent is taking part in both events, contributing an evidence score equal to the premised referential hypothesis' evidence score. If the exact same scene graph node takes part in two events, an exact matching existent value, \textit{r}, is used as the evidence score. An example of a causal hypothesis can be seen in Figure \ref{figure:set-17-generation-examples}b. 

\paragraph{Affective Hypothesis Generation}

Affective relationships concern the emotions, motivations, and goals of the characters in the images. They provide a human experiential component beyond purely observational information, which most humans can naturally understand and relate to. The system currently implements hypothesis generation for each ConceptNet relationship categorized as an affective relationship. 

Affective hypothesis generation is premised on existing knowledge evidence. Existing knowledge evidence is found by looking at the scene graph for events that existents identified as people or animals take part in. That a character takes part in an event suggests that they might feel a certain way or have some motivation or goal, and that these affective evaluations were formed either before or after taking part in the event. From the concept node of an event that a character takes part in, the system searching for edges in ConceptNet with relationships categorized as affective. The terminating concepts of those affective relationships are then assigned with affective relationships to the character that took part in the event. Evidence scores are the weights of the affective edges in ConceptNet. An example of a affective hypotheses and their evidence can be seen in Figure \ref{figure:set-17-generation-examples}c. 

\subsubsection{Hypothesis Evaluation Process}
\label{section:hypothesis_evaluation}

Of the over-generated set of hypotheses, only a limited number will be kept. In sensemaking, while speculation does lead to richer and better-interconnected representations of the world, such speculation must be kept internally consistent, rather than allowing wild speculation which is contradictory or without basis \citep{weick2005organizing}. 

As the task of selecting the set of facts to accept in the system's knowledge representation requires the system to consider the trade-offs between several possibly conflicting objectives at once, we treat the task as a multi-objective optimization problem (MOP). MOPs are standard problem formulations that appear in various fields of science, engineering, and economics \citep{deb2014multi}. 

\paragraph{MOP Formulation for Hypothesis Evaluation}

Following the general formulation of MOPs, we define the problem as finding the hypothesis set or sets, $h_m$, with the highest objective score for a set of objective functions, $f_i(x)$, where each hypothesis set is part of the set of feasible hypothesis sets $H_f$:

\begin{equation}
    \label{equation:moo_equation_hypothesis}
    max_1^m(f_i(x)) | h_j \in H_f
\end{equation}

Three objective functions will be used - connectivity ($f_1$), density ($f_2$), and support ($f_3$). Connectivity and density are closely related, and are both measures of graph topology. Support is a measure of the evidence quantified in the hypothesis generation step. 
\paragraph{Connectivity} is the minimum number of nodes that must be removed from a graph for it to be separated into independent subgraphs. High connectivity promotes relationships across sections of the knowledge graph which are sparsely interconnected, such as those formed from individual scene graphs for each image, and between stranded or sparsely connected individual nodes and the rest of the graph. 

\paragraph{Density} is the proportion of edges in the graph versus the maximum number of edges possible. High density promotes drawing relationships between as many concepts as possible. Density can be calculated in linear time. 

\paragraph{Support} is the degree to which the hypotheses accepted by the system are supported by their evidence. For the system to accept hypotheses it is more sure about, stronger evidence should be promoted. Support is calculated as the sum of the evidence scores for each hypothesis in a hypothesis set. 

\paragraph{Constraints}

As mentioned previously, each hypothesis set $h_m$ is a member of the set of feasible hypothesis sets $H_f$. A feasible hypothesis set is one which satisfies each constraint $c_j$:

\begin{equation}
    \label{equation:constraint_equation_general}
    h \in H_f \iff \forall c_j, c_j(h) = true
\end{equation}

The constraints that will be used to check feasibility will be based on a set of toggleable heuristics-based checks. Currently, two contradiction checks have been implemented for referential hypotheses. This first is non-redundancy ($c_1$); a referential hypothesis cannot assert that the same existent appears in the same image in two different places at the same time. The second is the transitive property ($c_2$); if it is hypothesized that existent a and existent b are both existent c, then it must also be so that existent a is existent b.

\section{Examples}
\label{section:examples}

The following examples were generated by the current system on two different image sets. For each example, ConceptNet paths are capped at a length of three. The first example demonstrates different methods of hypothesis generation. The second demonstrates scoring for alternate valid hypothesis sets. 

\subsection{Example 1}
\label{section:example-1}

\begin{figure}[t]
\vskip 0.05in
\begin{center}
\includegraphics[width=0.6\textwidth]{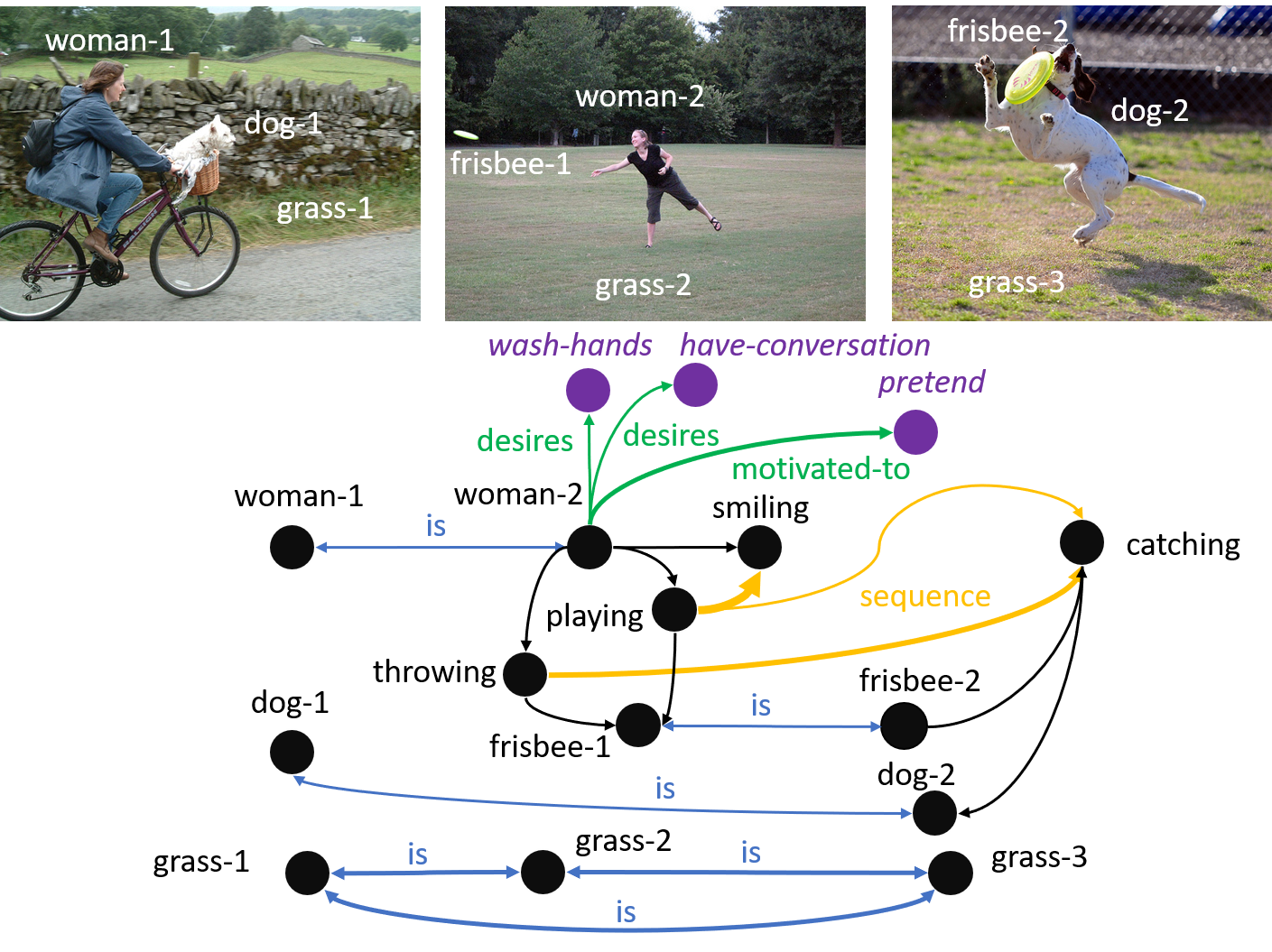}
\caption{Example image sequence with region labels (top) and excerpt of its system generated knowledge graph (bottom), showing scene graph nodes and edges (black), concept-net concepts (purple), and referential (blue), causal (orange), and affective (green) hypothesized relationships. Thickness of hypothesized relationship lines corresponds to evidence score of hypothesis.}
\label{figure:set-17-full-example}
\end{center}
\vskip -0.2in
\end{figure}

Figure \ref{figure:set-17-full-example} shows an image sequence and an excerpt of a knowledge graph generated by the sensemaking subsystem, with referential, causal, and affective hypotheses. An exact matching existent value of \textit{r=3} is used, equivalent to a referential hypothesis between two existents that have two matching attributes. 

\begin{figure}[t]
\vskip 0.05in
\begin{center}
\includegraphics[width=0.85\textwidth]{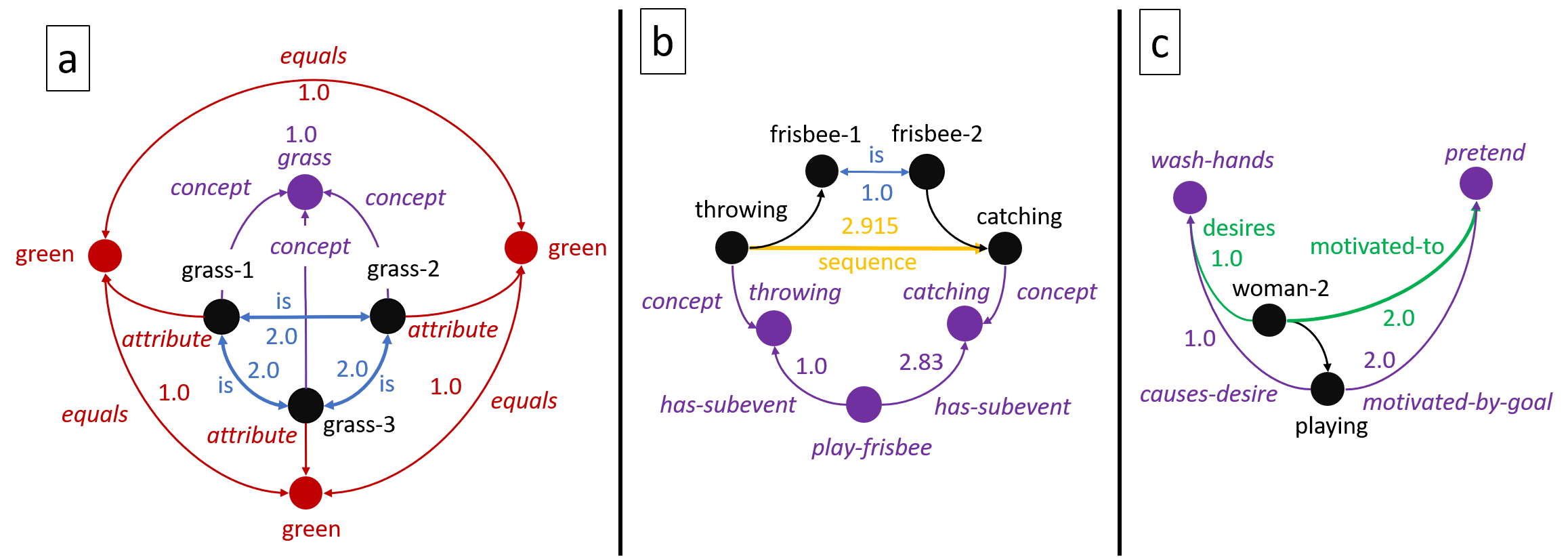}
\caption{Examples of referential (a, blue), causal (b, yellow), and affective (c, green) hypotheses from the knowledge graph in Figure \ref{figure:set-17-full-example} with their evidence with evidence scores, including observational evidence (red) and external knowledge evidence (purple). }
\label{figure:set-17-generation-examples}
\end{center}
\vskip -0.2in
\end{figure}

\paragraph{Referential Hypothesis Example}

Figure \ref{figure:set-17-generation-examples}a shows three referential hypotheses from Figure \ref{figure:set-17-full-example} and their evidence. The hypotheses are that all three observed 'grass' areas are actually instances of the same area of grass across all three images. The observational evidence for these hypotheses consists of the 'green' attribute found on all three scene graph nodes, each of which was only included by a single annotator and thus contributes a score of 1 to each hypothesis' total evidence score. The existing knowledge evidence for these hypotheses is that the 'grass' scene graph nodes correspond to the same ConceptNet concept node for \textit{grass}, as seen in the purple \textit{concept} links in Figure \ref{figure:set-17-generation-examples}a, contributing an evidence score of 1. Summing the observational and existing knowledge evidence scores yields each hypothesis' total evidence score of 2.

As the existing knowledge evidence for referential hypotheses can never be greater than 1, referential hypotheses are more susceptible to observational differences and similarities than other hypotheses. This is especially true for attributes that carry high confidence (e.g. if three annotators had each added the 'green' attribute to each piece of grass) or for existents that have many different shared attributes (e.g. if all the grass was 'green', 'tall', and 'thin'). Though the sensemaking system relies more on its observations to premise referential hypotheses, because other hypotheses can be premised on referential hypotheses, the decisions involved in assigning referential hypotheses can still reach beyond visual similarities. This will be explored more in Section \ref{section:example-2}.

\paragraph{Causal Hypothesis Example}

Figure \ref{figure:set-17-generation-examples}b shows a single causal hypothesis from the knowledge graph in Figure \ref{figure:set-17-full-example} and its evidence. The hypothesis is that the event of the woman throwing the frisbee in image 2 is related enough to the event of the dog catching the frisbee in image 3 to be considered part of the same line of action. The existing knowledge evidence for the hypothesis in Figure \ref{figure:set-17-generation-examples}b is the path in ConceptNet between the two events' ConceptNet concept nodes - that throwing is a subevent of playing frisbee and that catching is a subevent of playing frisbee). The average weight of the edges of the path, 1.915, is the evidence's score. 

The hypothesis in Figure \ref{figure:set-17-generation-examples}b is premised on an existing hypothesis. According to a prior referential hypothesis, the frisbee that takes part in the throwing event is the same as the one in the catching event. As such, the evidence score of the \textit{is} referential hypothesis that states that the frisbees are the same is added to the evidence score of the causal hypothesis, for a total evidence score of 2.915. 

The system considers any two events that either have a shared participating existent or have a causal path in ConceptNet as possible parts of a causal sequence. With shared participating existents, the strength of the evidence depends on the strength of its prior referential hypothesis. This means that the system is more confident about causal hypotheses between events when the system is more confident they are about the same existents. As in Weick's model for sensemaking, past inferences bolster subsequent inferences.

For causal paths in ConceptNet, the strength of evidence depends on the average weight of the path. Though all ConceptNet paths were capped at a length of three for these examples, longer, less related paths are more likely to contribute lower confidence evidence than shorter paths traversing higher weight edges. Conceptually, reaching for longer, more tangential topical connections results in more tenuous conclusions than shorter, more apparent or more trustworthy topical connections. 

\paragraph{Affective Hypothesis Example}

Figure \ref{figure:set-17-generation-examples}c shows a pair of affective hypotheses from Figure \ref{figure:set-17-full-example} and their evidence. The hypothesis is that the woman in the second image wants to wash their hands and has a goal of playing pretend. This stems from the 'playing' event that the woman in the second image takes part in. In ConceptNet, the affective relationships from the \textit{playing} concept are that playing is motivated by the goal of playing pretend and that playing causes a desire to wash one's hands. The resulting affective hypotheses reflect this on the woman who is playing. The evidence scores are the weights of the affective edges in ConceptNet. 

The hypotheses displayed in Figure \ref{figure:set-17-full-example} display conclusions outside of the directly observed information in the images. Conclusions such as that two frisbees are the same, that observed events are connected, and that characters in the images have affective responses are external to the observed information, though they are related to, and even premised on, observational information. 

\subsection{Example 2}
\label{section:example-2}

\begin{figure}[t]
\vskip 0.05in
\begin{center}
\includegraphics[width=0.6\textwidth]{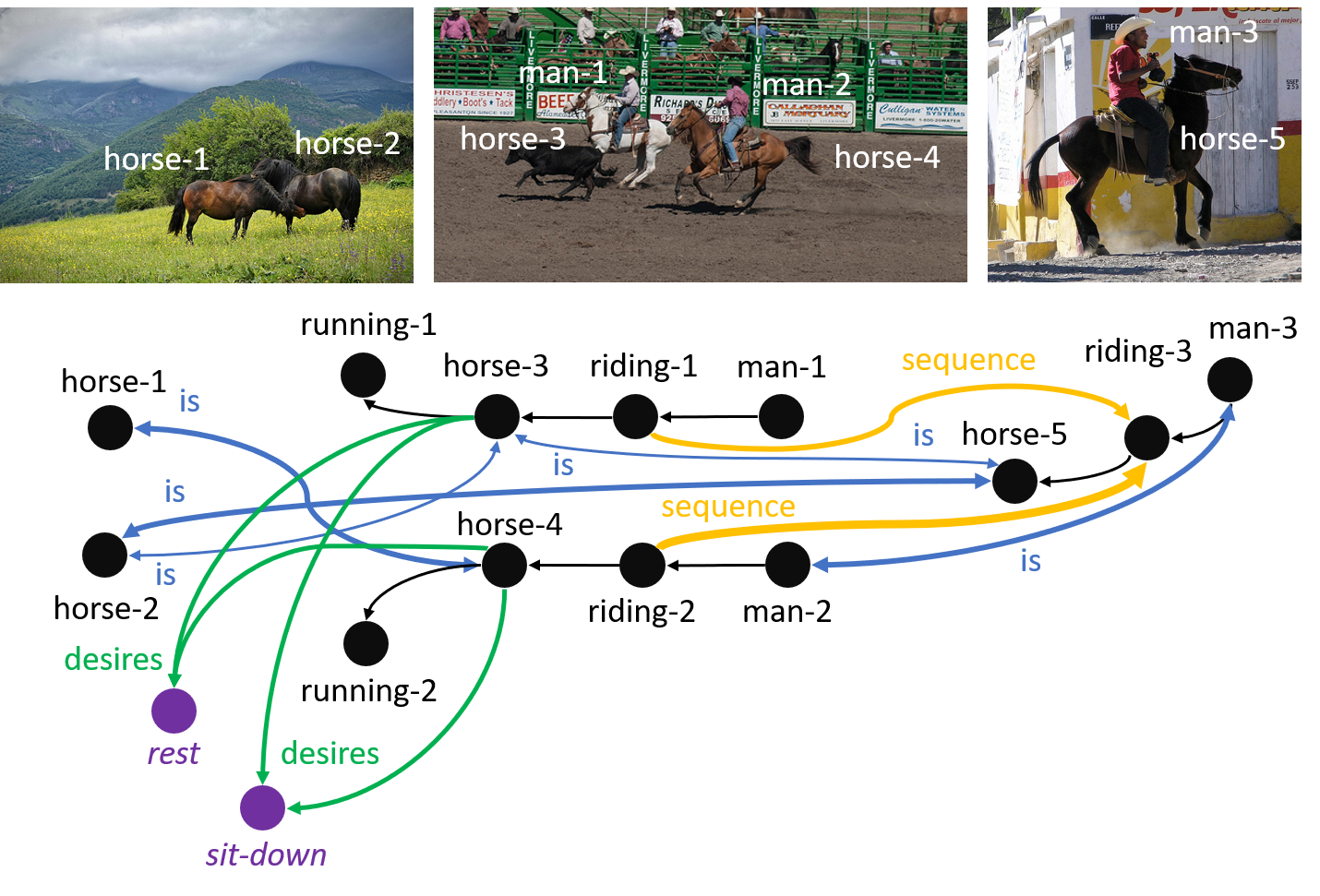}
\caption{Another example of an image sequence with region labels and an excerpt of its system generated knowledge graph, as in Figure \ref{figure:set-17-full-example}.}
\label{figure:set-28-full-example}
\end{center}
\vskip -0.2in
\end{figure}

The example in Figure \ref{figure:set-28-full-example} shows a series of images with multiple contradicting referential hypotheses, stemming from the multiple instances of 'horse' and 'man' existents within the images. As discussed in Section \ref{section:hypothesis_evaluation}, referential \textit{is} hypotheses must obey the transitive property. Thus, if horse-2 \textit{is} horse-3 and horse-2 \textit{is} horse-5, if the hypothesis horse-3 \textit{is} horse-5 is not included in the hypothesis set, there is a contradiction and the hypothesis set is rejected. Similarly, \textit{is} hypotheses cannot be placed between existents in the same image. Thus, accepting the hypothesis horse-1 \textit{is} horse-2 or horse-3 \textit{is} horse-4 leads to the hypothesis set being rejected.

\begin{figure}[t]
\vskip 0.05in
\begin{center}
\includegraphics[width=0.6\textwidth]{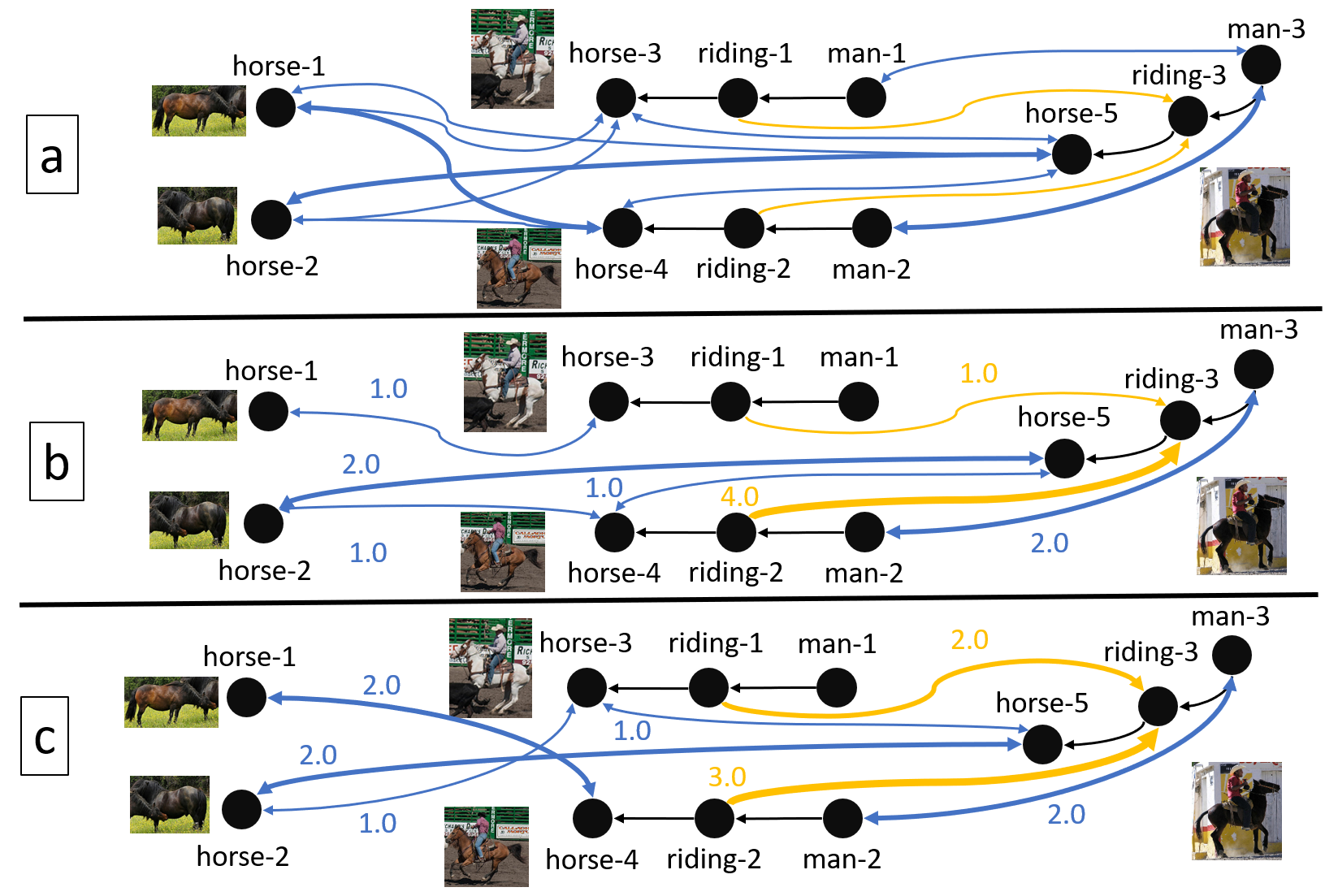}
\caption{The full set of possible referential and causal relationships for an excerpt of the knowledge graph in Figure \ref{figure:set-28-full-example} (a) and two possible hypothesis sets (b, c) from different resolutions of contradicting hypotheses.}
\label{figure:set-28-example}
\end{center}
\vskip -0.2in
\end{figure}

This leads to several possible hypothesis sets with different \textit{is} hypotheses. Figure \ref{figure:set-28-example}a shows all the possible referential hypotheses in blue (excluding hypotheses between existents in the same image) and two possible causal hypotheses in orange. All referential hypotheses are between existents with the same concept, with different evidence strength based on visual attribute matching. For this example, color is the only matching attribute. 

Both causal hypotheses are between the two 'riding' events in the second image and the single 'riding' event in the third image, representing the supposition that these events lie on the same line of action (e.g. the men riding horses during the second image somehow leads to, or is related to, the man riding a horse in the last image). The strength of both causal hypotheses is based in part on the referential hypotheses they are premised on, and as such is based on how the \textit{is} hypotheses are resolved. Without any referential hypotheses, their evidence scores are the same: the score of 1 for the 0-length path in ConceptNet through their shared 'riding' concept. 

Figure \ref{figure:set-28-example}b and c show two different resolutions of \textit{is} hypotheses, each representing a different hypothesis set. The density and connectivity scores of both hypothesis sets are the same. As such, we will only be discussing their support scores (the sum of their evidence scores) when discussing the resolution of hypothesis sets. 

In Figure \ref{figure:set-28-full-example}b, the hypothesis set results in a single causal hypothesis with a high evidence score. The black horse in the first image is equated to the brown horse in the second image and the black horse in the final image. This allows the man in the red shirt to be the same man in both the second and last images, which leads to a stronger causal hypothesis; though the color of his horse does not match between images 2 and 3, the system asserts that the man in the red shirt is still the same red-shirted rider of the same horse as before. As mentioned in Section \ref{section:hypothesis_generation}, stronger observational evidence leads to stronger causal hypotheses. The causal connection between riding-2 and riding-3 is premised on more certain referential hypotheses: the man taking part in both riding events is wearing a red shirt. The support score for this hypothesis set is 12.

The hypothesis set in Figure \ref{figure:set-28-full-example}c results in a balance between observational evidence and causal connection surety. Unlike the previous hypothesis, it asserts that the man in red and the horse in the final image were not riding together in the second image. Instead, the man in red switched to riding the horse which was previously being ridden by the man in white. This makes both causal connections to riding-3 have at least one returning existent: horse-3 from the riding-1 event and man-2 from the riding-2 event. Thus, while neither causal connection is as strong as the one between riding-2 and riding-3 in the hypothesis set in Figure \ref{figure:set-28-full-example}b, together, they exhibit the same total causal evidence strength. Additionally, though rider and horse change, all of the observational evidence that can match is matched. The support score for this hypothesis set is 13, making it the better candidate of the two.

What we can observe from this example is that the system can sacrifice maximal surety in some hypotheses for overall better surety of the whole knowledge graph. As in human sensemaking, it is not always the singularly most accurate conclusion that makes the most sense, but the one that makes an individual's overall understanding of a situation more internally cohesive and confident. Overall, more of the information is also interconnected in the second hypothesis set versus the first. In the first, one of the horse and rider pairs in the second image is entirely unconnected to the events in the final image. In the second, one horse and one rider from each pair becomes a recurring character, and there is more reason for both horse and rider pairs to appear in the second image. 

\section{Conclusion}

In this paper, we have explored two important components of how human beings understand the world - the process of sensemaking and the connections of narrative. Both are successful and ubiquitous for humans, and both would be of benefit for computational systems that also seek to understand the world. We have described one domain of computational systems that would benefit from sensemaking and narrative - the visual storytelling task - as well as our plans and progress for a computational system for the visual storytelling task that uses both the sensemaking processs and the connections inherent in narrative for the purposes of creating a representation that can improve story generation for images beyond the directly observable. 

We are examining future work in three directions: implementation of non-sensemaking subsystems, evaluation, and including additional relationships and types of coherence relationships.

Currently, the focus of the project is on the sensemaking subsystem, with the Visual Genome dataset serving as a proxy for a functional computer vision subsystem and the output of the sensemaking subsystem being considered as the output of the current system. A full system would include a computer vision subsystem that utilizes existing scene graph generation methods, such as Graph-RCNN \citep{yang2018graph}, and narrative generation from prior work \citep{battad2016using,battad2019facilitating} would be implemented to turn the sensemaking subsystem's knowledge graph output into a text story that can be directly compared to the outputs of other visual storytelling systems. 

Relatedly, a more comprehensive evaluation of the system's current output, while difficult because of the difference between a knowledge graph and the text passages generated by other visual storytelling systems, would be beneficial. In the interim, the terms and relationships introduced by the system into its knowledge graph atop its base scene graph can be compared to those used by human writers writing stories for the same images, e.g. those used in the Visual Genome dataset, versus those written by human annotators for region descriptions. 

Finally, the specific relationships implemented within each of the three coherence relationship types do not represent every relationship that could be implemented within each type, but were considered sufficient in the current system to explore their implementation and interactions with each other and within the system. Exploring other specific referential (such as part-of or member-of), causal, and affective relationships can lead to a wider interconnection of nodes and more layered hypothesis generation (e.g. more hypotheses premised on other hypotheses). Temporal and spatial relationships can also be implemented a similarly limited way to the current relationships implemented. Finally, conflict checking within hypothesis evaluation currently only exists for referential relationships. Similar heuristics for each type of coherence relationship would have to be designed for each one implemented.

 
\begin{acknowledgements} 
\noindent
We would like to thank our colleagues in the Cognitive Science Department and in the Cognitive and Immersive Systems Lab (CISL) at RPI, with special thanks to Dr. Cara Reedy for their invaluable feedback. 
\end{acknowledgements}

\vspace{-0.25in}

{\parindent -10pt\leftskip 10pt\noindent
\bibliographystyle{cogsysapa}
\bibliography{bibliography.bib}

}


\end{document}